\documentclass[letterpaper, 10 pt, conference]{ieeeconf}  

\IEEEoverridecommandlockouts                              

\usepackage{amsmath} 
\usepackage{amssymb}  
\usepackage{multirow}
\usepackage{array}
\usepackage{booktabs}
\usepackage{graphicx}
\usepackage{algorithm}
\usepackage{algpseudocode} 
\usepackage{xcolor}
\usepackage{caption}
\usepackage{hyperref}
\usepackage[export]{adjustbox}

\title{\LARGE \bf
CaT: Constraints as Terminations for Legged Locomotion Reinforcement Learning
}

\author{Elliot Chane-Sane$^{*}$$^{1}$, Pierre-Alexandre Leziart$^{*}$$^{1}$, Thomas Flayols$^{1}$, \\ Olivier Stasse$^{1,2}$, Philippe Sou\`eres$^{1}$, Nicolas Mansard$^{1,2}$
\thanks{$^{*}$Equal contribution}
\thanks{$^{1}$LAAS-CNRS, Universit\'e de Toulouse, Toulouse, 31400, France
        {\tt\small first.last@laas.fr}}
\thanks{$^{2}$Artificial and Natural Intelligence Toulouse Institute, Toulouse, France.}%
}

\begin{document}

\makeatletter
\let\@oldmaketitle\@maketitle
\renewcommand{\@maketitle}{\@oldmaketitle
  \centering \adjincludegraphics[trim={0 0 0 {0.15\height}}, clip,width=1.0\linewidth]
  {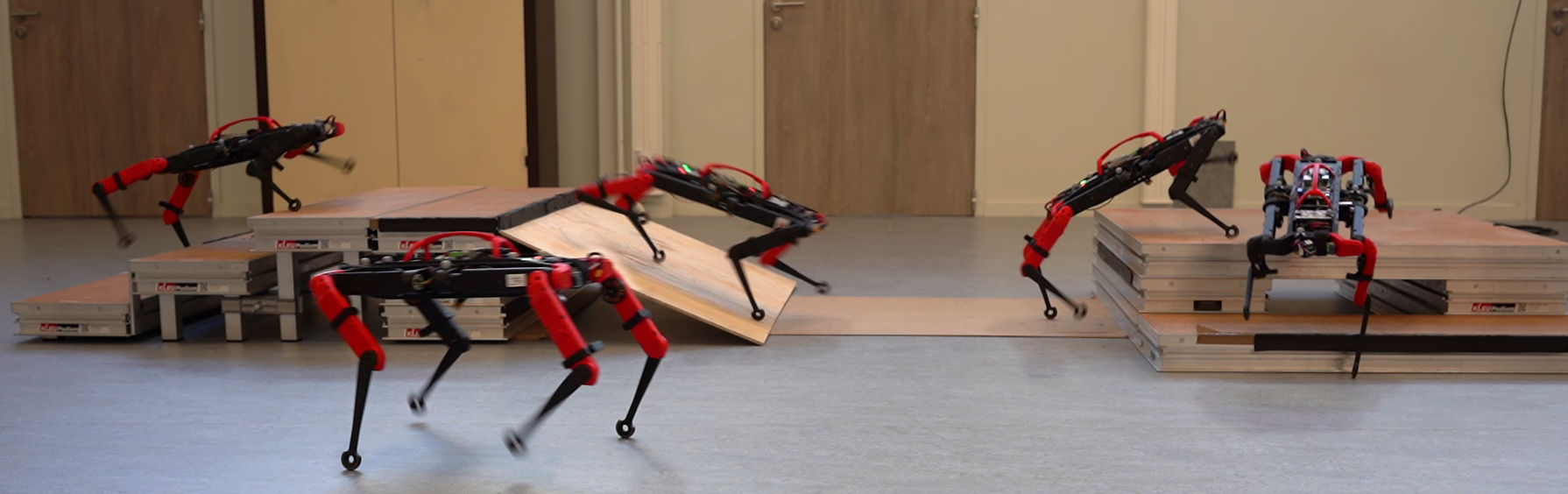} 
    \captionof{figure}{
    The open-hardware quadruped robot Solo-12 trained with CaT performing agile locomotion over challenging terrains while satisfying safety and style constraints.
    The robot can walk up stairs, traverse slopes, and climb over high obstacles.
    } \label{fig:teaser}  }
\makeatother

\maketitle
\thispagestyle{empty}
\pagestyle{empty}


\begin{abstract}
Deep Reinforcement Learning (RL) has demonstrated impressive results in solving complex robotic  tasks such as quadruped locomotion.
Yet, current solvers fail to produce efficient policies respecting hard constraints.
In this work, we advocate for integrating constraints into robot learning and present \textit{Constraints as Terminations (CaT)}, a novel constrained RL algorithm.
Departing from classical constrained RL formulations, we reformulate constraints through stochastic terminations during policy learning: any violation of a constraint triggers a probability of terminating potential future rewards the RL agent could attain.
We propose an algorithmic approach to this formulation, by minimally modifying widely used off-the-shelf RL algorithms in robot learning (such as Proximal Policy Optimization).
Our approach leads to excellent constraint adherence without introducing undue complexity and computational overhead, thus mitigating barriers to broader adoption.
Through empirical evaluation on the real quadruped robot Solo crossing challenging obstacles, we demonstrate that CaT provides a compelling solution for incorporating constraints into RL frameworks.
Videos and code are available at \href{https://constraints-as-terminations.github.io}{constraints-as-terminations.github.io}.
\end{abstract}

\section{Introduction}

Deep reinforcement learning (RL) has proven highly effective in crafting control policies for complex robotic tasks.
In quadruped locomotion, RL approaches have demonstrated high performances to train policies capable of traversing challenging terrains~\cite{agarwal2023legged, cheng2023parkour, hoeller2023anymal, zhuang2023robot} and generating natural, animal-like motions~\cite{peng2020learning, escontrela2022adversarial, li2023learning}.
In this work, we follow recent successful approaches based on model-free RL~\cite{schulman2017proximal} to train policies on a curriculum of increasingly difficult settings~\cite{bengio2009curriculum, soviany2022curriculum} in simulation and directly transfer the learned policy on the physical robot~\cite{rudin2022learning, chen2023learning, bellegarda2022visual} to overcome challenging obstacles.
Compared to previous approaches in robot motion~\cite{kajita2003biped,farshidian2017efficient,leziart2021implementation}, this workflow requires minimal design choices, relying on generic algorithms and simulations that allow to generate a wide variety of tasks.

Yet, reward shaping remains a meticulous endeavor as it demands a delicate balance between accomplishing the desired task, adhering to physical limitations, enabling seamless sim-to-real transfer, and ensuring natural and efficient motions.
Many of these terms could be more effectively and intuitively formulated as constraints.
For instance, joint torque and velocity limits have clear physical meanings that should not be considered through a hyperparameter search.
While incorporating such constraints aligns with common practices in model-based control~\cite{dantec2022whole, risbourg2022real, leziart2022improved}, widespread adoption in robot learning has been limited.
Although some recent constrained RL methods have been applied to locomotion \cite{kim2023not, lee2023evaluation}, they often simplify reward engineering at the cost of algorithmic complexity, as additional critic networks and terms in the policy loss function have to be implemented.

In this work, we propose \textit{Constraints as Terminations (CaT)}, a streamlined approach for constrained RL that prioritizes simplicity and flexibility.
We introduce constraints through stochastic terminations during policy learning: any violation of a constraint leads to a probability of terminating the future rewards the RL agent could have achieved.
To do so, we down-scale all the future rewards based on the magnitude of the constraint violations during policy learning through the discount factor.
This naturally encourages the agent towards satisfying the constraints to maximize future rewards, while providing an alternative reward signal to recover from constraint violations.
This principle can be seen as a refined extension of the common practice of using a straightforward termination function, leveraging stochastic termination to yield a dense feedback to the policy.

Our approach is simple to implement and seamlessly integrates with existing off-the-shelf RL algorithms.
In our experiments, we instantiate CaT with Proximal Policy Optimization~\cite{schulman2017proximal} (PPO), a model-free on-policy algorithm widely used in robot learning.
We design a set of constraints to ensure that the learned policy can be safely deployed to the real robot, and a set of style constraints to exhibit natural motions.
We demonstrate the effectiveness of our approach by deploying locomotion policies on a Solo quadruped robot with height-scan observations, producing agile locomotion skills capable of traversing challenging terrains composed of stairs, a steep slope and a high platform (see Fig.~\ref{fig:teaser}).

In summary, our contributions are the following:
\begin{enumerate}
\item we introduce stochastic terminations as a way to shape the behavior of the policy to satisfy constraints in a minimalist fashion,
\item we propose constraint designs to enforce safe behaviors and make the policy adhere to a specific walking style on flat terrains, while letting RL adapt the style on rougher terrains,
\item and we validate our approach on a real Solo quadruped robot to overcome diverse obstacles in a parkour while satisfying safety and style constraints.
\end{enumerate}
\section{Related Work}

Reinforcement learning has emerged as a particularly effective method for obtaining agile and adaptive policies for quadruped robots.
While some approaches attempt to train RL locomotion policies directly on physical quadruped robots by leveraging sample-efficient RL techniques~\cite{smith2022walk, wu2023daydreamer}, a popular approach entails training policies in simulation before transferring them to the real world~\cite{peng2018sim, margolisyang2022rapid, aractingi2023controlling, aractingi2023hierarchical}. 
This transfer relies on accurate physics simulators and domain randomization to ensure policy transferability to the physical robot~\cite{Tan18SimtoReal, kumar2021rma, xie2021dynamics}. 
Recently, GPU-based simulators capable of simulating thousands of robots in parallel~\cite{todorov2012mujoco, brax2021github, makoviychuk2021isaac} have streamlined this process~\cite{rudin2022learning}. 
The resulting policies exhibit natural, animal-like motions and can adapt to challenging terrain configurations~\cite{fu2021minimizing, bellegarda2022robust, cheng2023parkour, agarwal2023legged, zhuang2023robot, hoeller2023anymal, duan2023learning}.
In our experiments, we follow this sim-to-real approach and deploy our policies on the Solo-12 robot~\cite{leziart2021implementation, grimminger2020open} for challenging terrain traversal.

Incorporating constraints is a common practice in model-based control, where their importance to ensure robot safety is commonly accepted~\cite{jallet2023proxddp, osqp, tonneau2020sl1m}.
Yet constraints have garnered limited attention in the RL community, where the main effective solvers do not readily consider them~\cite{schulman2017proximal, haarnoja2018soft} and achieving policies that comply with constraints is often done through intricate reward shaping.
In legged locomotion, this approach typically results in reward functions comprising numerous terms that are labor-intensive to tune.
For instance, the reward functions used in \cite{rudin2022learning, aractingi2023controlling} comprise a dozen of terms.
Moreover, the resulting policy, being a compromise among maximizing each of these terms, is not guaranteed to satisfy constraints in all situations~\cite{lee2023evaluation}.

Prior works have explored the imposition of constraints or safety mechanisms in addition to rewards within the learning process to ensure safety guarantees.
Recovery policies have been learned jointly with the locomotion policy to address safety violations~\cite{yang2022safe, he2024agile}.
\cite{alshiekh2018safe, fan2024learn} proposed to shield the learning agent by directly substituting policy actions by safe actions whenever necessary to prevent constraint violations.
Other approaches incorporate constraint satisfaction directly into the policy optimization algorithms by adjusting the policy update rules to discourage violations.
For instance, Lagrangian methods~\cite{chow2018risk,tessler2018reward} approach constrained problems as unconstrained ones by introducing Lagrange multipliers, but this often leads to instability due to hyperparameter sensitivity~\cite{achiam2017constrained}.
More closely related to our work, \cite{kim2023not} modifies the Interior-point Policy Optimization algorithm \cite{liu2020ipo} and demonstrate quadruped locomotion skills on rough-terrain whereas
\cite{lee2023evaluation} implements a modified Penalized Proximal Policy Optimization (P3O) \cite{zhang2022penalized} algorithm on a wheeled quadruped robot, both showcasing enhanced safety in the learned policies and facilitating the tuning of reward terms at the cost of additional algorithmic complexity.
By contrast, our approach is simple to implement, requiring minimal changes to existing locomotion RL pipelines and introducing no additional computational overhead.

Terminating the future rewards and resetting the episode is ubiquitously used in reinforcement learning to avoid certain behaviors.
For instance, \cite{rudin2022learning} terminates the episode  with a low reward when the robot base or knees touch the ground.
\cite{sun2022constrained} further showed that learning policies for early-terminated Markov decision processes (ET-MDP), i.e. terminating future rewards on constraint violations without necessarily resetting the environment, is an effective way to learn constraint-satisfying policies.
However, our experiments highlight that this approach does not readily scale to complex systems such as quadruped robots with dozens of constraints.
We propose in the next section to capitalize on this common practice to design a novel approach to enforce generic hard constraints in RL. 
To that end, we first reformulate the constraint as a probability of satisfaction. 
Then we introduce stochastic terminations as a way to downscale the sum of future possible rewards while keeping a dense feedback to the policy, in particular by keeping informative direction from the domain outside constraint satisfaction.

\section{Method}

\begin{figure*}[ht]
\centering
\adjincludegraphics[trim={0 {0.24\height} 0 {0.04\height}},clip,width=0.5\linewidth]{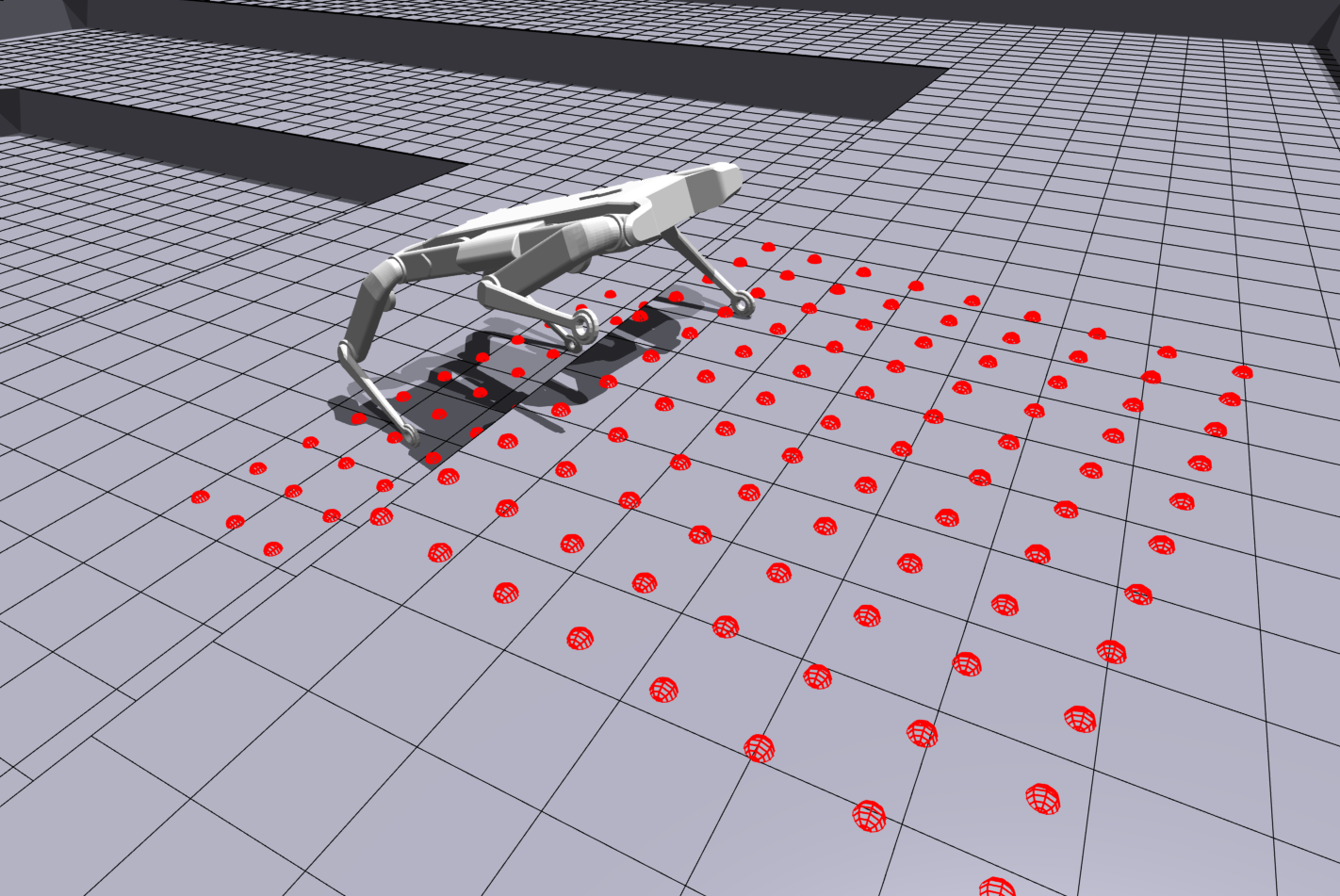}\hfill 
\adjincludegraphics[trim={0 {0.14\height} 0 {0.14\height}},clip,width=0.5\linewidth]{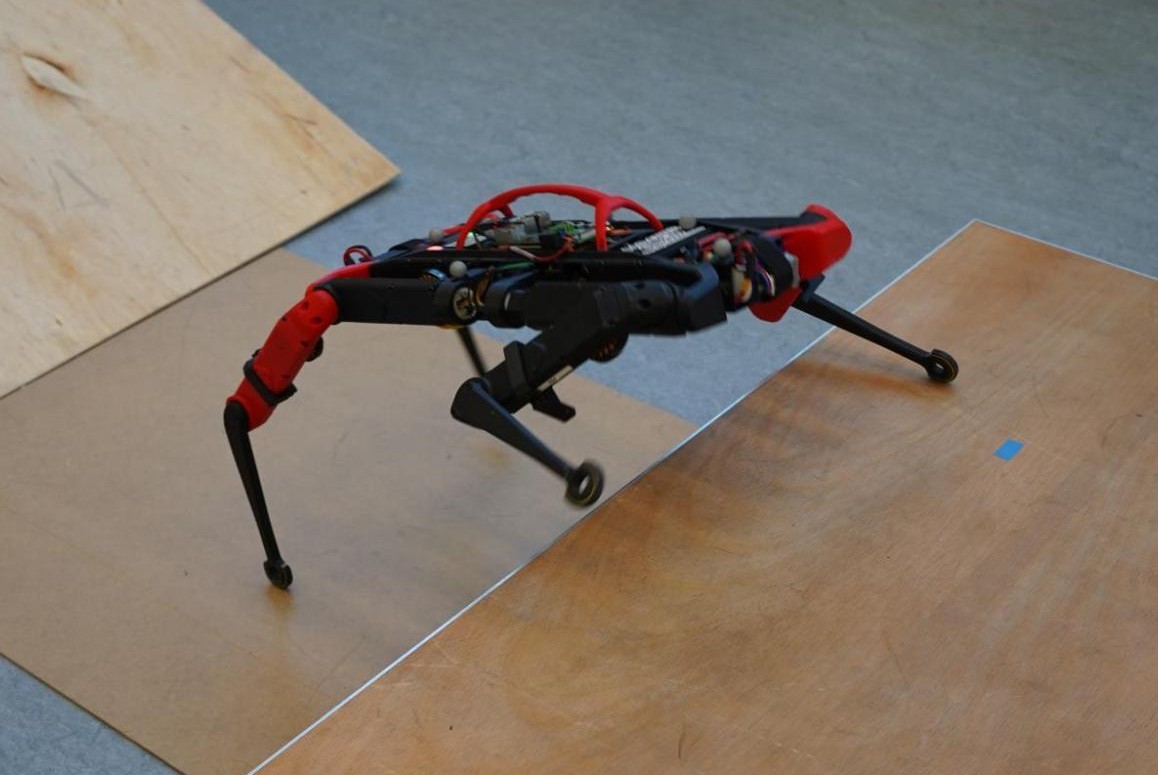}
\caption{
(Left) The quadruped robot is trained with CaT in simulation using height-map scan.
(Right) The learned policy is directly deployed on the real robot.
Knowing the obstacle course on which the robot is placed, we use external motion capture cameras to reconstruct the height-map of its surroundings based on its position and orientation in the world.
}
\label{fig:simtoreal}
\vspace{-0.5cm}
\end{figure*}

\subsection{Problem Formulation}

We consider an infinite, discounted Markov Decision Process $\mathcal{S}, \mathcal{A}, r, \gamma, \mathcal{T}$ with state space $\mathcal{S}$, action space $\mathcal{A}$, reward function $r$, discount factor $\gamma$ and dynamics $\mathcal{T}$.
RL aims to find a policy $\pi$ that maximizes the discounted sum of future rewards:
\begin{equation}
\max_\pi \mathbb{E}_{\tau \sim \pi, \mathcal{T}} \left [ \sum_{t=0}^\infty \gamma^t r(s_t, a_t) \right ].
\label{eq:mdp}
\end{equation}
In the following, we assume positive rewards $r \geqslant 0$ for simplicity (without loss of generality w.r.t. any other lower bounded definition).
Constrained RL additionally introduces a set of constraint functions $\{ c_i: \mathcal{S} \times \mathcal{A} \rightarrow \mathcal{R}, i \in I\}$ and aims to maximize rewards while limiting the discounted sum of constraints over the trajectories generated by the policy: 
\begin{equation}
    \mathbb{E}_{\tau \sim \pi, \mathcal{T}} \left [ \sum_{t=0}^\infty \gamma^t c_i(s_t, a_t) \right ] \leq \epsilon_i \text{ } \forall i \in I.
\end{equation}
While standard in the RL literature \cite{achiam2017constrained, kim2023not}, this formulation includes a notion of budget for the constraints.
We consider instead maximizing rewards while avoiding constraint violation at each time step:
$\mathbb{E}_{\tau \sim \pi, \mathcal{T}} \left [ \sum_{t=0}^\infty \gamma^t 1_{ c_i(s_t, a_t) > 0} \right ] \leq \epsilon_i,$
where $1_{\gamma^t c_i(s_t, a_t) > 0}$ indicates whether the $i$-th constraint has been violated at time $t$.
This is equivalent to:
\begin{equation}
\mathbb{P}_{(s, a) \sim \rho^{\pi, \mathcal{T}}_\gamma} \left [ c_i(s, a) > 0 \right ] \leq \tilde{\epsilon_i} \text{ } \forall i \in I,
\label{eq:cstr_prob}
\end{equation}
where $\rho^{\pi, \mathcal{T}}_\gamma$ corresponds to the discounted state-action occupancy distribution of the policy $\pi$.
While this corresponds to a special case of the more general constrained RL setting, this formulation, akin to chance-constrained optimization~\cite{charnes1959chance,nemirovski2007convex}, encompasses many practical applications of RL for robotic control.

\subsection{Constraints as Terminations}

\subsubsection{Reformulation}
Instead of directly solving (\ref{eq:mdp}) under the constraints (\ref{eq:cstr_prob}), we propose to reformulate it as:
\begin{equation}
\max_\pi \underset{\tau \sim \pi}{\mathbb{E}} \! \left [ \sum_{t=0}^\infty \! \left ( \prod_{t'=0}^t \gamma (1 - \delta (s_{t'}, a_{t'})) \! \right ) \! r(s_t, a_t) \right ], 
\label{eq:cat_mdp}
\end{equation}
where we introduce a random variable $\delta_t : \mathcal{S} \times \mathcal{A} \rightarrow [0, 1]$  indicating whether the episode terminates and the future rewards are terminated from time step $t$.
Importantly, we propose to design $\delta_t$ as a function of the constraint violations $c_i$.
Note that episode terminations are \textbf{not} environment resets, but merely future reward terminations from a policy learning perspective.
Under the expectation, the Bernoulli variable and its probability are the same.
In the rest of the paper, $\delta_t$ will refer directly to the probability of termination.

\subsubsection{Naive termination}
A naive approach is to terminate the future rewards if any constraint is violated~\cite{sun2022constrained} with the following binary function for $\delta$:
\begin{equation}
    \delta = 1 - \prod_{i \in I} 1_{c_i \leq 0}.
    \label{eq:etmdp}
\end{equation}
\cite{sun2022constrained} showed that if the minimum value of the rewards is high enough, which can be easily obtained by adding a high enough constant value, the learned policy will satisfy the constraints.
However, terminating the episode if any constraint is violated might be overly conservative with respect to the constraints and impair exploration and learning.
Moreover, such a termination condition offers a sparse signal for recovering from constraint violations: once the agent enters a region of constraint violation, the episode always terminates and the agent does not learn anything.

\subsubsection{Stochastic terminations}
We propose that $\delta_t$ can take values beyond $0$ or $1$ depending on the constraint violations at time $t$.
As a result, any violation of a constraint leads to a probability of terminating the future rewards the RL agent could have achieved.
If no constraints are violated, then the episode terminates with a probability of zero, whereas if one or more constraints are violated, $\delta$ may take positive values between $0$ and $1$.
In that case, the sum of all future rewards at $t$ and after time are re-scaled by $(1-\delta_t)$.
Therefore, in order to maximize the sum of future rewards, the agent naturally gravitates towards satisfying the constraints.
Allowing $\delta$ to take values in $]0, 1[$ enables the agent to learn to recover from constraint violations.
Moreover, depending on the value of $\delta$, this allows some exploration inside the region of constraint violation.

By designing $\delta$ such that it is increasing with $c_i$, the termination probability will provide a dense signal to the learning algorithm to recover from constraints.
Driven by simplicity, we propose the following termination probability function:
\begin{equation}
\label{eq:delta_function}
\delta = \max_{i \in I} \: p_i^\text{max} \text{clip}(\frac{c_i^+}{c_i^\text{max}}, 0, 1),
\end{equation}
where $c_i^+ = \max(0, c_i(s, a))$ is the violation of constraint $i$, $c_i^\text{max}$ is an exponential moving average of the maximum constraint violation over the last batch of experience collected in the environment:
\begin{equation}
c_i^\text{max} \gets \tau^c c_i^\text{max} + (1 - \tau^c) \max_{(s, a) \in \text{batch}} c_i^+(s, a),
\end{equation}
with decay rate $\tau^c \in ]0, 1[$ and $p_i^\text{max}$ a hyperparameter that controls the maximum termination probability for the constraint $i$.
We found directly using the maximum over the batch of experience without exponential moving average to be slightly less stable.
Hence, the termination probability for each constraint is proportional to the magnitude of the constraint violation, while the dynamic update of $c_i^\text{max}$ makes sure that the termination function always provides a relevant learning signal throughout training.
We found that this design was simple to implement while achieving effective constraint satisfaction.

\begin{algorithm}
\caption{Implementation of CaT with PPO, with alterations from the original RL algorithm highlighted in \textcolor{red}{red}.}
\label{alg:CaT}
\begin{algorithmic}[1]
\For{$\text{epoch} = 1$ \textbf{to} $N$}
    \State data $\gets$ PPO.collect\_trajectories()
    \State \textcolor{red}{compute $\delta(\text{data.constraints})$ using (\ref{eq:delta_function})}
    \State \textcolor{red}{$\text{data.rewards} \gets \text{data.rewards} \times (1-\delta)$}
    \State \textcolor{red}{$\text{data.dones} \gets \delta$}
    \State PPO.update\_policy(data)
\EndFor
\end{algorithmic}
\end{algorithm}
\vspace{-0.3cm}

Our proposed approach, \textit{Constraints as Terminations (CaT)}, can easily be incorporated into existing RL algorithms with minimal changes, by simply computing $\delta$ based on the constraint violations using (\ref{eq:delta_function}), multiplying the rewards by $\delta$ and rewriting the terminations with $\delta$.
These modifications can be implemented with just a few lines of codes to existing RL algorithms.
Algorithm \ref{alg:CaT} highlights the changes needed to implement our approach on top of PPO.


\section{Application to Legged Locomotion}

We train a policy in simulation using CaT and directly transfer the policy to a real Solo-12 robot (see Fig.~\ref{fig:simtoreal}).
For this quadruped locomotion problem, the state space $\mathcal{S}$ corresponds to the measured positions $q_t$ and velocities $\dot{q_t}$ of all 12 joints of the robot, the previous action $a_{t-1}$ and the linear and angular velocity commands $v^\text{des}_{xy}$ and $\omega^\text{des}_z$ that the robot must track.
For non-blind navigation, the robot also observes the height-scan $h_\text{scan}$ of its surroundings.
The action space $\mathcal{A}$ corresponds to desired joint position offsets $a_t = \Delta q^\text{des}_t$ with respect to a default joint configuration $q^{\star}$, that are then converted to torques through a proportional-derivative (PD) controller operating at a higher frequency than the neural policy.
The derivative part of the controller aims to bring the joint velocity to zero.

\begin{table}[tbp]
\vspace{0.15cm}
\centering
\begin{tabular}{c|c}
\toprule
\multicolumn{2}{c}{Task formulation: through rewards (Option A)} \\
\midrule
Reward function & $r = e^{-\frac{\left \| v^\text{des}_{xy} - v_{xy} \right \|^2_2 }{0.25}} + \frac{1}{2} e^{-\frac{\left | \omega^\text{des}_z - \omega_z \right |^2}{0.25} }$ \\
\bottomrule
\toprule
\multicolumn{2}{c}{Task formulation: through soft constraints (Option B)} \\
\midrule
Reward function & $r=1$\\
Linear velocity tracking & $c_\text{lin vel} = \left \| v^\text{des}_{xy} - v_{xy} \right \|_2 - \epsilon_\text{track}$ \\
Angular velocity tracking & $c_\text{ang vel} = \left | \omega^\text{des}_{z} - \omega_{z} \right | - \epsilon_\text{track}$ \\
\bottomrule
\toprule
\multicolumn{2}{c}{Hard constraints for safety} \\
\midrule
Knee or base collision & $c_\text{knee/base contact} = 1_\text{knee/base contact}$ \\
Foot contact force & $c_{\text{foot contact}_j} = \|f^{\text{foot}_j}\|_2 - f^\text{lim}$ \\
\bottomrule
\toprule
\multicolumn{2}{c}{Soft constraints for safety ($\forall k \in 1..12$)} \\
\midrule
Torque limits & $c_{\text{torque}_k} = |\tau_k| - \tau^\text{lim}$ \\
Joint velocity limits & $c_{\text{joint velocity}_k}= |\dot{q_k}| - \dot{q}^\text{lim}$ \\
Joint acceleration limits & $c_{\text{joint acceleration}_k}= |\ddot{q_k}| - \ddot{q}^\text{lim}$ \\
Action rate limits & $c_{\text{action rate}_k}= \frac{\left | \Delta q^\text{des}_{t, k} - \Delta q^\text{des}_{t-1, k} \right |}{dt} - \dot{q}^\text{des lim}$  \\
\bottomrule
\toprule
\multicolumn{2}{c}{Soft constraints for style} \\
\multicolumn{2}{c}{(Active on flat terrains only, $\forall j \in 1..4)$} \\
\midrule
Base orientation & $c_\text{ori} = \left \| \text{base ori}_{\text{xy}} \right \|_2 - \text{base}^\text{lim}$ \\
Hip orientation & $c_{\text{hip}_j} = |\text{hip ori}_j| - \text{hip}^\text{lim}$ \\
Foot air time & $c_{\text{air time}_j} = t_\text{air time}^\text{des} - t_{\text{air time}_j}$ \\
Number of foot contacts & $c_\text{n foot contacts} = |n_\text{foot contact} - n_\text{foot contact}^\text{des}|$ \\
Stand still if $v^\text{des}=0$ & $c_{\text{still}}= (\|q - q^\star \|_2 - \epsilon_\text{still} ) \times 1_{v^\text{des} = 0}$ \\

\bottomrule
\end{tabular}
\caption{Rewards and constraints used in our experiments.}
\label{table:constraint_list}
\vspace{-0.8cm}
\end{table}

One might consider that each reward and constraint can serve one of these three purposes:
\begin{itemize}
    \item define the task to be achieved,
    \item ensure that the generated trajectories are safe and transferable to the physical robot,
    \item or impose a style to the generated motions.
\end{itemize}
The complete list of rewards and constraints used in our experiments is provided in Table~\ref{table:constraint_list}.
We detail them below.

\paragraph{Task definition}
The legged locomotion task is to track the linear velocity command in horizontal direction $v^\text{des}_{xy}$ and yaw rate $\omega^\text{des}_z$.
We consider a velocity tracking reward function widely used in RL for legged locomotion (Option A)~\cite{rudin2022learning, lee2023evaluation}.
Alternatively, we propose to define the velocity tracking task as a constraint to be satisfied (Option B).

\paragraph{Safety constraints}
Safety constraints are defined to ensure the policy learned in simulation will transfer well and safely to the physical robot once training is complete.
We prohibit collisions to the knee and the base of the robot to avoid dangerous behaviors that might destroy the robot.
We limit the contact force of each foot $n$ to prevent the robot from hitting the ground too harshly, and we limit the torque applied to each joint $k$ to prevent damaging the actuators.
To ensure that the generated motions are smooth for seamless sim-to-real transfer, we also limit joint velocities, joint accelerations and action rates.

\paragraph{Style constraints}
Style constraints are used to guide learning towards natural-looking motions.
However, defining relevant style constraints in any terrain configuration is difficult.
We propose to enforce style constraints only on flat surfaces while deactivating them (i.e. set them to $0$) otherwise.
This allows us to define a precise style to follow on flat terrains while providing room for the RL algorithm to adapt the learned behavior on more challenging terrains.
In our implementation, the terrain is considered flat if the variance of the scan dots is below a certain threshold $\text{var}(h_\text{scan}) < \text{var}_\text{scan}^\text{lim}$.
We limit the orientation of the base and the angle of the hips.
When the velocity command is above a threshold, we additionally ground the flying phase duration of each foot and limit to two the number of foot contacts with the ground whereas, if no velocity is provided, we force the robot to go back to its default pose.

\paragraph{Soft and hard constraints}
Our method introduces a hyperparameter $p_i^\text{max}$ for each constraint which trades off exploration with constraint satisfaction.
A high value of $p_i^\text{max}$ will ensure that the constraint is strictly satisfied during training but might lead to overly conservative exploration, whereas lower values of $p_i^\text{max}$ will allow the learning agent to discover higher reward regions of the behavior space.
Motivated by simplicity, we propose to classify constraints into two groups: \textit{hard constraints} with $p_i^\text{max}=1.0$ for constraints that should never be violated, and \textit{soft constraints}, where $p_i^\text{max}$ increases from $0.05$ to $0.25$ throughout the course of training, that the RL algorithm might violate during exploration and learn to recover from.
We found that this design allowed the agent to maximally learn complex locomotion skills while further enforcing the constraints in the later stage of training.
In our experiments, base or knee contact collisions and foot contact forces are defined as hard constraints and the rest of the constraints as soft ones.

This set of constraints results in a large constraint vector comprising more than $60$ terms.
While prior approaches group constraints together~\cite{kim2023not, lee2023evaluation}, we found that this additional engineering burden was unnecessary for CaT.

\begin{table}[tbp]
\vspace{0.15cm}
\centering
\begin{tabular}{c|cc}
\toprule
Method & Rewards & Cstr.\\
\midrule
Hard constraints only & 0 & $n.a.$\\
ET-MDP~\cite{sun2022constrained} & 0 & $n.a.$ \\
N-P3O~\cite{zhang2022penalized,kim2023not,lee2023evaluation} & 593.2 ($\pm$ 49.5)  & 8\% ($\pm$ 1\%)\\
CaT (Tracking Rewards) & \textbf{682.9} ($\pm$ 5.8) & \textbf{0.5\%} ($\pm$ 0.3\%)\\
\bottomrule
\end{tabular}
\caption{
Average sum of rewards (\textit{Rewards}) and average time proportion of torque constraint violation for any joint (\textit{Cstr.}) achieved by the policies on flat terrain in simulation.
Results are averaged over 4 training seeds.
}
\label{table:results_simu}
\vspace{-0.7cm}
\end{table}

\begin{table*}[tbp]
\vspace{0.15cm}
\centering
\begin{tabular}{c|cc|cc|cc|cc|cc}
\toprule
\multirow{2}{*}{Method}  & \multicolumn{2}{c|}{Front Stairs} & \multicolumn{2}{c|}{Sideways Stairs} & \multicolumn{2}{c|}{Slope} & \multicolumn{2}{c|}{Platform} & \multicolumn{2}{c}{Average}\\
 & Succ. & Cstr. & Succ. & Cstr. & Succ. & Cstr. & Succ. & Cstr. & Succ. & Cstr.  \\
\midrule
Style always active & 50.0\% & 2.5\% & 40.0\% & 4.8\% & 30.0\% & 2.0\% & 17.5\% & 5.4\% & 34.4\% & 3.7\% \\
CaT (Tracking Rewards) & \textbf{100.0\%} & \textbf{0.08\%} & 42.5\% & \textbf{0.3\%} & \textbf{97.5\%} & \textbf{0.3}\% & 77.5\% & \textbf{1.1\%} & 79.4\% & \textbf{0.5\%} \\
CaT (Tracking Constraints) & \textbf{97.5\%} & 0.5\% & \textbf{85.0\%} & 1.8\% & \textbf{95.0\%} & 1.2\% & \textbf{85.0\%} & 3.4\% & \textbf{90.6\%} & 1.7\% \\
\bottomrule
\end{tabular}
\caption{
Average success rate (\textit{Succ.}) and average time proportion of torque constraint violation for any joint (\textit{Cstr.}) achieved by the policies on the different obstacles of the parkour on the real robot:
walking up the stairs from the front (\textit{Front Stairs}) and sideways (\textit{Sideways Stairs}), walking up the slope (\textit{Slope}) and walking up the platform as high as the robot's base (\textit{Platform}).
Results are averaged over $4$ random training seeds and $10$ attempts per obstacle per seed.
}
\label{table:results}
\vspace{-0.5cm}
\end{table*}

\section{Experiments}

\subsection{Experimental setup}

To train our policies, we leverage the PPO algorithm~\cite{schulman2017proximal} using the implementation from rl-games~\cite{rl-games2021}, which we slightly modified to accommodate non-boolean terminations, alongside massively parallel simulation of Isaac Gym~\cite{makoviychuk2021isaac}.
Hyperparameters are provided in Appendix \ref{appendix:hyperparameters}.
Blind policies for flat terrains are trained for 2000 epochs whereas policies with height-scan map are trained for 20000 epochs for agile terrain traversal.
This amounts to respectively 1 hour and 10 hours of training on a single V100 GPU.
Except for CaT specific implementations, the resulting training procedure is similar to~\cite{rudin2022learning}.

After training in simulation, the controller is directly deployed on a real Solo-12 robot.
The policy runs at 50 Hz on a Raspberry Pi 4 Model B using a custom C++ implementation.
Target joint positions are sent to the onboard PD controller running at 10 kHz.
PD gains are kept low to obtain a compliant impedance controller that will achieve a behavior close to torque control and will be able to dampen and absorb impacts \cite{aractingi2023controlling}. This is further made possible thanks to the transparent actuation of Solo-12.
For more details on the hardware, please refer to \cite{leziart2021implementation, grimminger2020open}.
Instead of directly capturing a height-scan map of the robot's surrounding terrain, we use motion capture to track the position of the robot and sample the corresponding height map points.

To validate the agility of the learned policies in diverse scenarios, we evaluate our approach on a challenging obstacle parkour comprising a set of stairs, a slope and a platform roughly the height of the robot (see Fig.~\ref{fig:teaser}).
Following Table~\ref{table:constraint_list}, we consider two versions of CaT: one with the task defined through rewards (\textit{CaT (Tracking Rewards)}) and one with the task defined through constraints (\textit{CaT (Tracking Constraints)}).
We compare CaT to the following baselines:
\begin{itemize}
    \item \textit{ET-MDP}: a modification of our method designed to resemble~\cite{sun2022constrained} by using (\ref{eq:etmdp}) to compute $\delta$.
    \item \textit{N-P3O}: our reproduction of P3O~\cite{zhang2022penalized} using techniques from~\cite{kim2023not, lee2023evaluation}.
    \item \textit{Hard constraints only}: an ablation of our approach where we use $p_i^\text{max}=1.0$ for all constraints.
    \item \textit{Style always active}: an ablation of our approach where style constraints are always enforced.
\end{itemize}
For N-P3O, we group constraints of the same type together following~\cite{kim2023not, lee2023evaluation}, use dense constraint functions as in CaT as opposed to indicator functions used in~\cite{kim2023not, lee2023evaluation}, and employ foot phase duration and number of foot contacts as rewards rather than constraints, as N-P3O struggles to converge otherwise.
Solo-12 is a light robot with dynamic, but limited actuators that should avoid applying a torque of more than $3$Nm.
To evaluate the capabilities of our approach to enforce constraints, we focus on the torque constraint satisfaction and report the proportion of time where this constraint is violated for one or more joints.

\subsection{Results and Analysis}

We first compare \textit{CaT (Tracking Rewards)} to N-P3O, ET-MDP and \textit{Hard constraints only} trained on a flat terrain for blind locomotion in simulation.
Table~\ref{table:results_simu} reports the rewards and the torque constraints satisfaction achieved by the policies.
ET-MDP entirely fails to learn locomotion policies in our high-dimensional constraint problem.
This may be due to the fact that at the beginning of training, the robot always violates some constraints, preventing any reward or constraint feedback to allow policy learning.
Similarly, when the constraints are enforced too roughly (\textit{Hard constraints only}), learning fails completely, as overly stringent enforcement of constraints hinders exploration and learning.
Despite being simpler, CaT outperforms N-P3O, in both the sum of tracking rewards attained and the satisfaction of torque constraints after 2000 epochs of training.
We hypothesize that the integration of rewards and constraints into a unified RL framework allows CaT to learn faster.

\begin{figure}[t!]
\centering
\adjincludegraphics[trim={0 0 {0.15\width} 0},clip,width=0.33333\linewidth]{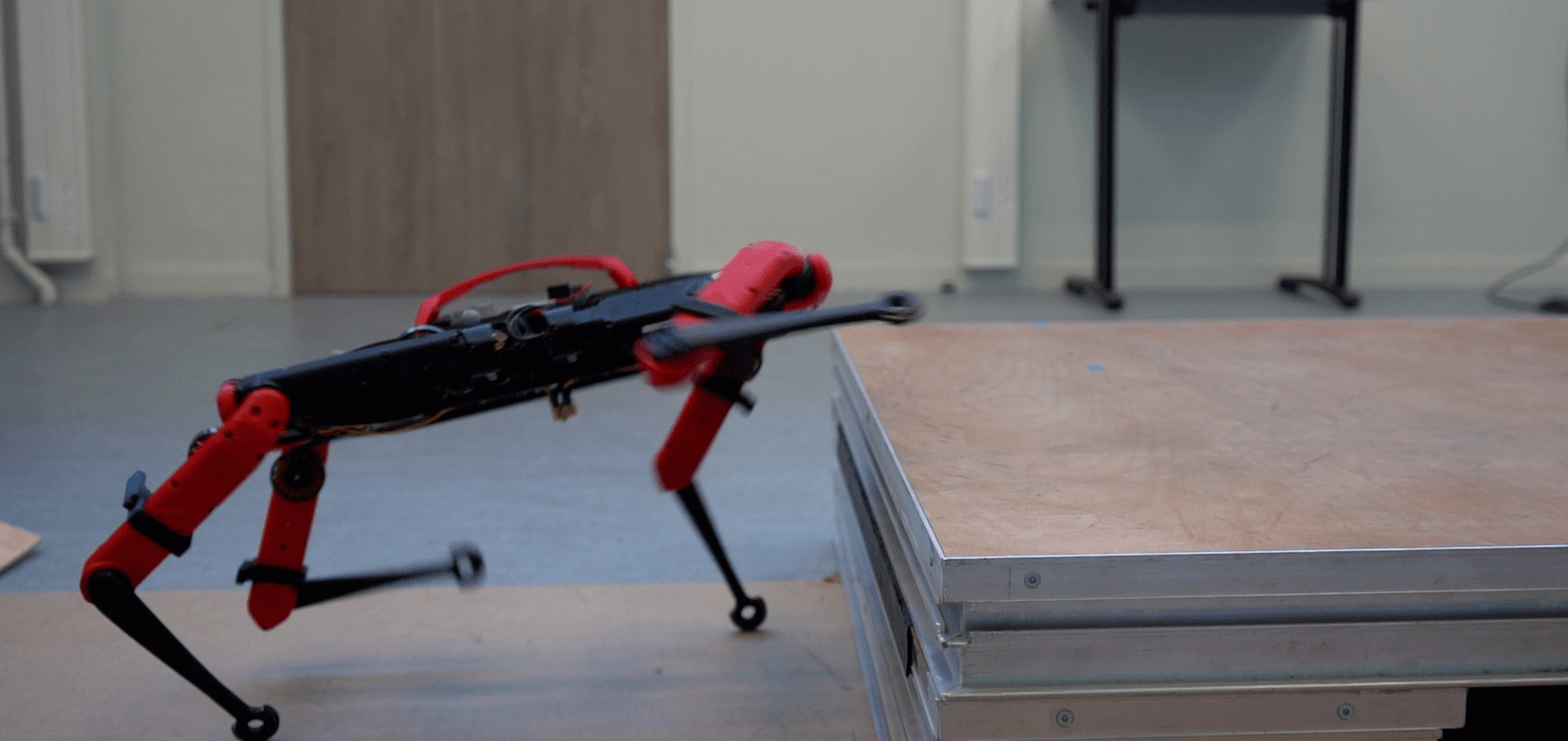}\hfill
\adjincludegraphics[trim={0 0 {0.15\width} 0},clip,width=0.33333\linewidth]{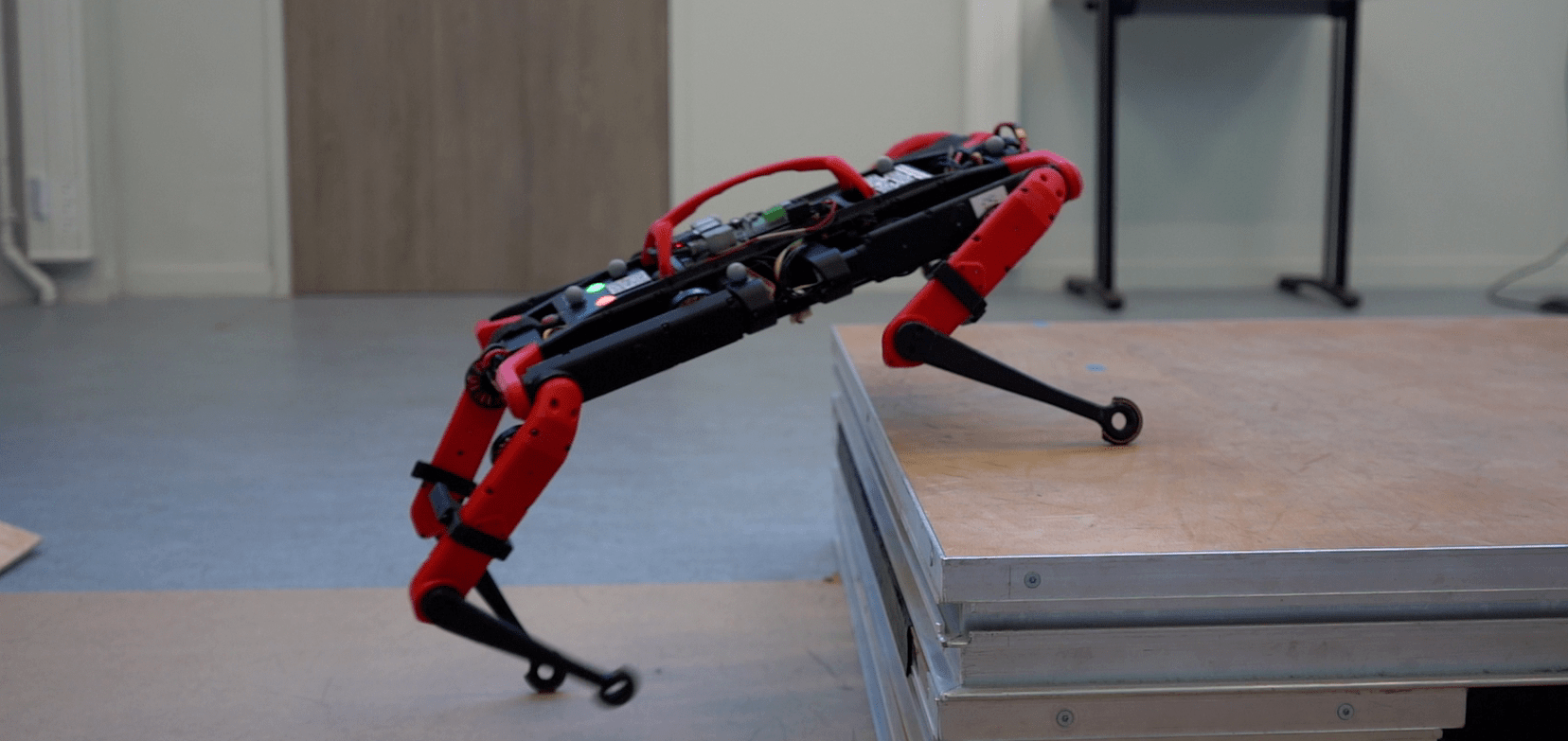}\hfill
\adjincludegraphics[trim={0 0 {0.15\width} 0},clip,width=0.33333\linewidth]{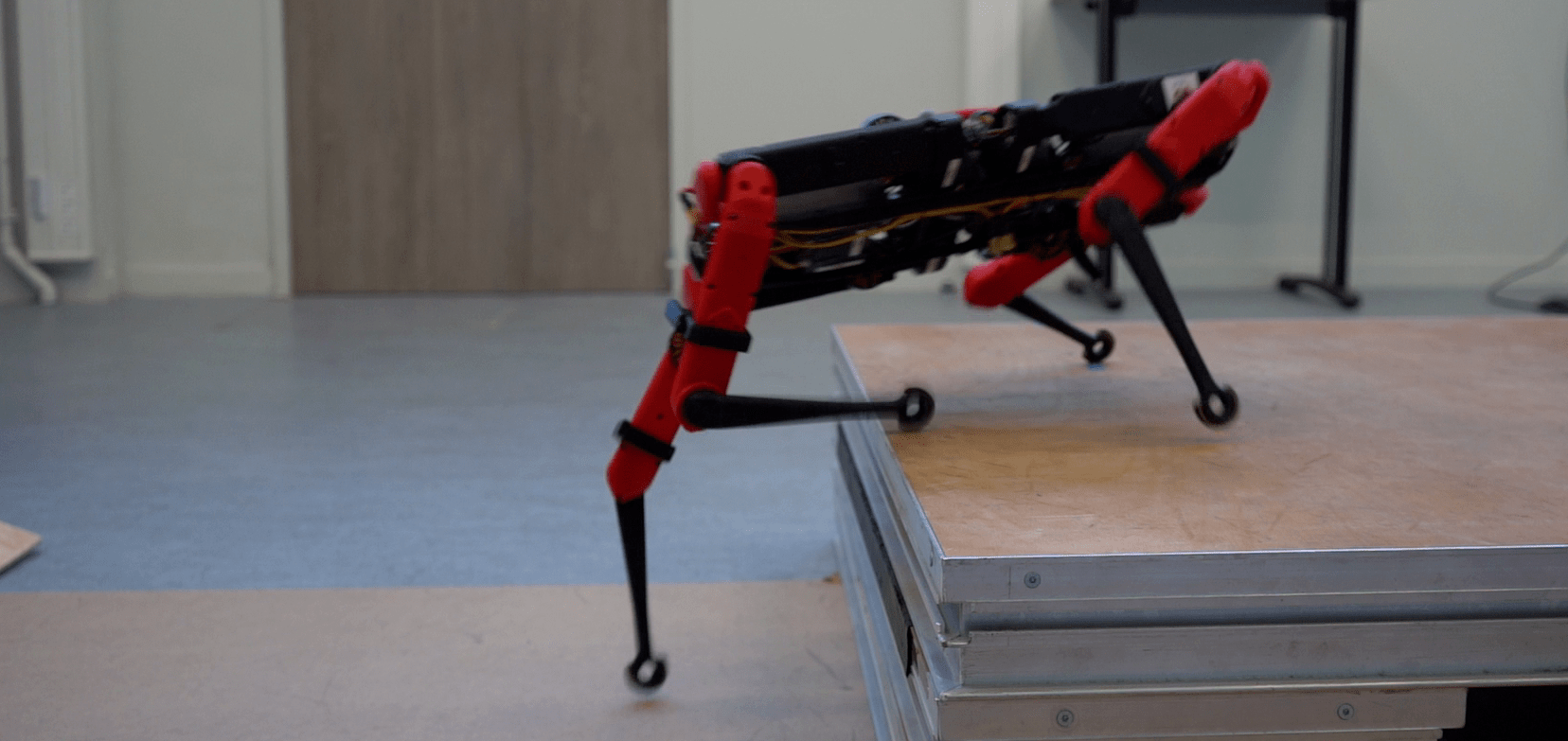}

\includegraphics[width=1.0\linewidth]{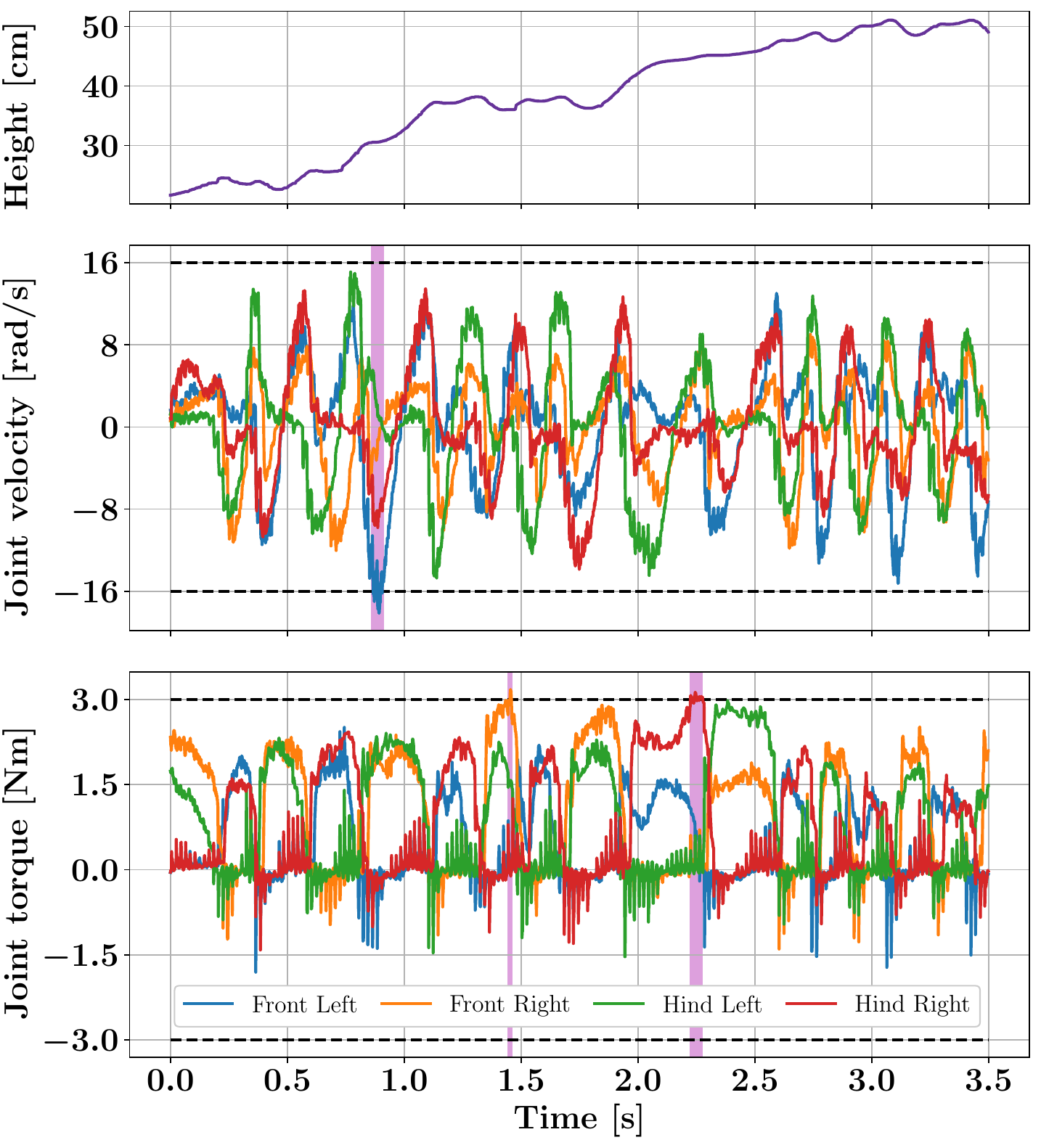}

\caption{Joint torques and velocities during the climb of a 24 cm platform. For clarity, we only report data for the knee joints, which had the highest torque peaks.}
\label{fig:torque}
\vspace{-0.8cm}
\end{figure}

Next, we deploy CaT with height-scan map on the real robot.
In Table~\ref{table:results}, we report the success rate of traversing each obstacle in the parkour.
CaT with both sets of rewards and constraints successfully learns agile locomotion skill to overcome each obstacle of the parkour.
Fig.~\ref{fig:teaser} shows a full traversal of the obstacle parkour, demonstrating natural motions on flat surfaces while achieving agile skills on more challenging obstacles.
Notably, CaT successfully learns to overcome all the obstacles while satisfying the torque constraint.
Fig.~\ref{fig:torque} shows that, while climbing on the platform almost as high as the robot, the torque remains within the limit set during training.
Interestingly, \textit{CaT (Tracking Constraints)}, where the locomotion task is defined entirely through constraints, learns agile locomotion skills.
In particular, it outperforms \textit{CaT (Tracking Rewards)} on climbing the stairs sideways, a difficult task where the height-scan map provides less visibility.
By contrast, \textit{CaT (Tracking Rewards)} often refuses to walk over the stairs sideways while achieving similar performances on other obstacles.
We hypothesize that \textit{CaT (Tracking Constraints)} is more prone to explore unsafe behaviors to fulfill the task constraints, resulting in better success rates at the expense of more constraint violations. 
This highlights how stochastic termination functions can be used to appropriately shape the behavior of the robot policy, either to ensure the controller is safe and adhere to a certain style, but also to fully define the intended task for the robot.

We then compare CaT to always enforcing style constraints, even on challenging terrains (\textit{Style always active}).
While this approach successfully learns walking skills on flat and rough terrains, it struggles on more difficult obstacles.
This occurs because adhering strictly to certain style constraints, as defined on flat surfaces, may not be compatible with other scenarios.
For example, imposing the constraint that the robot's base must remain horizontal is incompatible with scenarios involving stair climbing.
This is particularly striking when attempting to climb the platform, which requires tilting the base and lift the shoulders, as illustrated in Fig. \ref{fig:torque} (top).

\begin{figure}[t]
\centering
\includegraphics[width=1.0\linewidth]{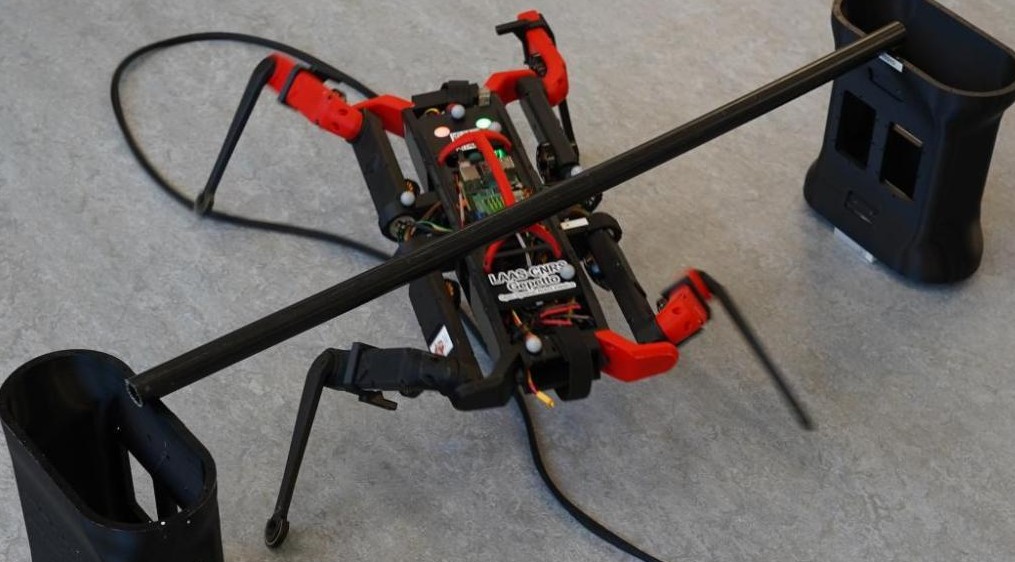}
\caption{
CaT trained with a constraint that limits the height of the base learns crouching locomotion skills.
}
\label{fig:crouching}
\vspace{-0.6cm}
\end{figure}

In Fig.~\ref{fig:crouching}, we illustrate how simply adding a constraint to limit the height of the base ($c_\text{height} = \text{height}_\text{base} - \text{height}^\text{max}_\text{base}$) can learn crouching locomotion skills on the quadruped.
Videos of the robot traversing the parkour and crouching are available in the supplementary video.
\section{Conclusion}

In this study, we introduce \textit{CaT}, a novel and minimalist algorithm addressing constraints in reinforcement learning.
We formulate the problem so that the probability of constraint violation is bounded and use stochastic termination to seamlessly integrate it on top of standard algorithms such as PPO.
On a Solo-12 quadruped robot, CaT successfully manages to learn agile locomotion skills on challenging terrain traversals, showcasing its utility in enforcing safety and stylistic constraints within quadruped locomotion.
Future work could explore more principled ways to define the termination conditions based on the constraints.

From a practical standpoint, constrained RL significantly simplifies the reward engineering process.
However, unlike previous, more intricate methods, our approach is notably simpler to implement, necessitating minimal code adjustments and is devoid of any computational overhead.
We hope the effectiveness and simplicity of our approach will foster the democratization of constrained RL in robotics.

\section*{Acknowledgements}
This work was funded in part by the COCOPIL project of R{\'e}gion Occitanie (France), the AS2 ANR-22-EXOD-0006 of the French PEPR O2R, the Dynamograde joint laboratory (grant ANR-21-LCV3-0002) and ROBOTEX 2.0 (Grants ROBOTEX ANR-10-EQPX-44-01 and TIRREX-ANR-21-ESRE-0015). It was granted access to the HPC resources of IDRIS under the allocations 2021-AD011012947 and 2023-AD011014301 made by GENCI.


\addtolength{\textheight}{-3cm}   

\bibliographystyle{IEEEtran}
\bibliography{biblio}

\begin{thebibliography}{10}
\providecommand{\url}[1]{#1}
\csname url@samestyle\endcsname
\providecommand{\newblock}{\relax}
\providecommand{\bibinfo}[2]{#2}
\providecommand{\BIBentrySTDinterwordspacing}{\spaceskip=0pt\relax}
\providecommand{\BIBentryALTinterwordstretchfactor}{4}
\providecommand{\BIBentryALTinterwordspacing}{\spaceskip=\fontdimen2\font plus
\BIBentryALTinterwordstretchfactor\fontdimen3\font minus \fontdimen4\font\relax}
\providecommand{\BIBforeignlanguage}[2]{{%
\expandafter\ifx\csname l@#1\endcsname\relax
\typeout{** WARNING: IEEEtran.bst: No hyphenation pattern has been}%
\typeout{** loaded for the language `#1'. Using the pattern for}%
\typeout{** the default language instead.}%
\else
\language=\csname l@#1\endcsname
\fi
#2}}
\providecommand{\BIBdecl}{\relax}
\BIBdecl

\bibitem{agarwal2023legged}
A.~Agarwal, A.~Kumar, J.~Malik, and D.~Pathak, ``Legged locomotion in challenging terrains using egocentric vision,'' in \emph{Conference on Robot Learning}.\hskip 1em plus 0.5em minus 0.4em\relax PMLR, 2023, pp. 403--415.

\bibitem{cheng2023parkour}
X.~Cheng, K.~Shi, A.~Agarwal, and D.~Pathak, ``Extreme parkour with legged robots,'' \emph{arXiv preprint arXiv:2309.14341}, 2023.

\bibitem{hoeller2023anymal}
D.~Hoeller, N.~Rudin, D.~Sako, and M.~Hutter, ``Anymal parkour: Learning agile navigation for quadrupedal robots,'' \emph{arXiv preprint arXiv:2306.14874}, 2023.

\bibitem{zhuang2023robot}
Z.~Zhuang, Z.~Fu, J.~Wang, C.~Atkeson, S.~Schwertfeger, C.~Finn, and H.~Zhao, ``Robot parkour learning,'' in \emph{Conference on Robot Learning ({CoRL})}, 2023.

\bibitem{peng2020learning}
X.~B. Peng, E.~Coumans, T.~Zhang, T.-W. Lee, J.~Tan, and S.~Levine, ``Learning agile robotic locomotion skills by imitating animals,'' \emph{arXiv preprint arXiv:2004.00784}, 2020.

\bibitem{escontrela2022adversarial}
A.~Escontrela, X.~B. Peng, W.~Yu, T.~Zhang, A.~Iscen, K.~Goldberg, and P.~Abbeel, ``Adversarial motion priors make good substitutes for complex reward functions,'' in \emph{2022 IEEE/RSJ International Conference on Intelligent Robots and Systems (IROS)}.\hskip 1em plus 0.5em minus 0.4em\relax IEEE, 2022, pp. 25--32.

\bibitem{li2023learning}
T.~Li, Y.~Zhang, C.~Zhang, Q.~Zhu, J.~Sheng, W.~Chi, C.~Zhou, and L.~Han, ``Learning terrain-adaptive locomotion with agile behaviors by imitating animals,'' in \emph{2023 IEEE/RSJ International Conference on Intelligent Robots and Systems (IROS)}.\hskip 1em plus 0.5em minus 0.4em\relax IEEE, 2023, pp. 339--345.

\bibitem{schulman2017proximal}
J.~Schulman, F.~Wolski, P.~Dhariwal, A.~Radford, and O.~Klimov, ``Proximal policy optimization algorithms,'' \emph{arXiv preprint arXiv:1707.06347}, 2017.

\bibitem{bengio2009curriculum}
Y.~Bengio, J.~Louradour, R.~Collobert, and J.~Weston, ``Curriculum learning,'' in \emph{Proceedings of the 26th annual international conference on machine learning}, 2009, pp. 41--48.

\bibitem{soviany2022curriculum}
P.~Soviany, R.~T. Ionescu, P.~Rota, and N.~Sebe, ``Curriculum learning: A survey,'' \emph{International Journal of Computer Vision}, vol. 130, no.~6, pp. 1526--1565, 2022.

\bibitem{rudin2022learning}
N.~Rudin, D.~Hoeller, P.~Reist, and M.~Hutter, ``Learning to walk in minutes using massively parallel deep reinforcement learning,'' in \emph{Conference on Robot Learning}, 2022.

\bibitem{chen2023learning}
S.~Chen, B.~Zhang, M.~W. Mueller, A.~Rai, and K.~Sreenath, ``Learning torque control for quadrupedal locomotion,'' in \emph{2023 IEEE-RAS 22nd International Conference on Humanoid Robots (Humanoids)}.\hskip 1em plus 0.5em minus 0.4em\relax IEEE, 2023, pp. 1--8.

\bibitem{bellegarda2022visual}
G.~Bellegarda and A.~Ijspeert, ``Visual cpg-rl: Learning central pattern generators for visually-guided quadruped navigation,'' \emph{arXiv preprint arXiv:2212.14400}, 2022.

\bibitem{kajita2003biped}
S.~Kajita, F.~Kanehiro, K.~Kaneko, K.~Fujiwara, K.~Harada, K.~Yokoi, and H.~Hirukawa, ``Biped walking pattern generation by using preview control of zero-moment point,'' in \emph{2003 IEEE international conference on robotics and automation (Cat. No. 03CH37422)}, vol.~2.\hskip 1em plus 0.5em minus 0.4em\relax IEEE, 2003, pp. 1620--1626.

\bibitem{farshidian2017efficient}
F.~Farshidian, M.~Neunert, A.~W. Winkler, G.~Rey, and J.~Buchli, ``An efficient optimal planning and control framework for quadrupedal locomotion,'' in \emph{2017 IEEE International Conference on Robotics and Automation (ICRA)}.\hskip 1em plus 0.5em minus 0.4em\relax IEEE, 2017, pp. 93--100.

\bibitem{leziart2021implementation}
P.-A. L{\'e}ziart, T.~Flayols, F.~Grimminger, N.~Mansard, and P.~Sou{\`e}res, ``Implementation of a reactive walking controller for the new open-hardware quadruped solo-12,'' in \emph{2021 IEEE International Conference on Robotics and Automation (ICRA)}.\hskip 1em plus 0.5em minus 0.4em\relax IEEE, 2021, pp. 5007--5013.

\bibitem{dantec2022whole}
E.~Dantec, M.~Naveau, P.~Fernbach, N.~Villa, G.~Saurel, O.~Stasse, M.~Taix, and N.~Mansard, ``Whole-body model predictive control for biped locomotion on a torque-controlled humanoid robot,'' in \emph{2022 IEEE-RAS 21st International Conference on Humanoid Robots (Humanoids)}.\hskip 1em plus 0.5em minus 0.4em\relax IEEE, 2022, pp. 638--644.

\bibitem{risbourg2022real}
F.~Risbourg, T.~Corb{\`e}res, P.-A. L{\'e}ziart, T.~Flayols, N.~Mansard, and S.~Tonneau, ``Real-time footstep planning and control of the solo quadruped robot in 3d environments,'' in \emph{2022 IEEE/RSJ International Conference on Intelligent Robots and Systems (IROS)}.\hskip 1em plus 0.5em minus 0.4em\relax IEEE, 2022, pp. 12\,950--12\,956.

\bibitem{leziart2022improved}
P.-A. L{\'e}ziart, T.~Corb{\`e}res, T.~Flayols, S.~Tonneau, N.~Mansard, and P.~Sou{\`e}res, ``Improved control scheme for the solo quadruped and experimental comparison of model predictive controllers,'' \emph{IEEE Robotics and Automation Letters}, vol.~7, no.~4, pp. 9945--9952, 2022.

\bibitem{kim2023not}
Y.~Kim, H.~Oh, J.~Lee, J.~Choi, G.~Ji, M.~Jung, D.~Youm, and J.~Hwangbo, ``Not only rewards but also constraints: Applications on legged robot locomotion,'' \emph{arXiv preprint arXiv:2308.12517}, 2023.

\bibitem{lee2023evaluation}
J.~Lee, L.~Schroth, V.~Klemm, M.~Bjelonic, A.~Reske, and M.~Hutter, ``Evaluation of constrained reinforcement learning algorithms for legged locomotion,'' \emph{arXiv preprint arXiv:2309.15430}, 2023.

\bibitem{smith2022walk}
L.~Smith, I.~Kostrikov, and S.~Levine, ``A walk in the park: Learning to walk in 20 minutes with model-free reinforcement learning,'' \emph{arXiv preprint arXiv:2208.07860}, 2022.

\bibitem{wu2023daydreamer}
P.~Wu, A.~Escontrela, D.~Hafner, P.~Abbeel, and K.~Goldberg, ``Daydreamer: World models for physical robot learning,'' in \emph{Conference on Robot Learning}.\hskip 1em plus 0.5em minus 0.4em\relax PMLR, 2023, pp. 2226--2240.

\bibitem{peng2018sim}
X.~B. Peng, M.~Andrychowicz, W.~Zaremba, and P.~Abbeel, ``Sim-to-real transfer of robotic control with dynamics randomization,'' in \emph{2018 IEEE international conference on robotics and automation (ICRA)}.\hskip 1em plus 0.5em minus 0.4em\relax IEEE, 2018, pp. 3803--3810.

\bibitem{margolisyang2022rapid}
G.~Margolis, G.~Yang, K.~Paigwar, T.~Chen, and P.~Agrawal, ``Rapid locomotion via reinforcement learning,'' in \emph{Robotics: Science and Systems}, 2022.

\bibitem{aractingi2023controlling}
M.~Aractingi, P.-A. L{\'e}ziart, T.~Flayols, J.~Perez, T.~Silander, and P.~Sou{\`e}res, ``Controlling the solo12 quadruped robot with deep reinforcement learning,'' \emph{scientific Reports}, vol.~13, no.~1, p. 11945, 2023.

\bibitem{aractingi2023hierarchical}
------, ``A hierarchical scheme for adapting learned quadruped locomotion,'' in \emph{2023 IEEE-RAS 22nd International Conference on Humanoid Robots (Humanoids)}.\hskip 1em plus 0.5em minus 0.4em\relax IEEE, 2023, pp. 1--8.

\bibitem{Tan18SimtoReal}
J.~Tan, T.~Zhang, E.~Coumans, A.~Iscen, Y.~Bai, D.~Hafner, S.~Bohez, and V.~Vanhoucke, ``Sim-to-real: Learning agile locomotion for quadruped robots,'' in \emph{Proceedings of Robotics: Science and Systems}, 2018.

\bibitem{kumar2021rma}
A.~Kumar, Z.~Fu, D.~Pathak, and J.~Malik, ``Rma: Rapid motor adaptation for legged robots,'' \emph{arXiv preprint arXiv:2107.04034}, 2021.

\bibitem{xie2021dynamics}
Z.~Xie, X.~Da, M.~Van~de Panne, B.~Babich, and A.~Garg, ``Dynamics randomization revisited: A case study for quadrupedal locomotion,'' in \emph{2021 IEEE International Conference on Robotics and Automation (ICRA)}.\hskip 1em plus 0.5em minus 0.4em\relax IEEE, 2021, pp. 4955--4961.

\bibitem{todorov2012mujoco}
E.~Todorov, T.~Erez, and Y.~Tassa, ``Mujoco: A physics engine for model-based control,'' in \emph{2012 IEEE/RSJ International Conference on Intelligent Robots and Systems}.\hskip 1em plus 0.5em minus 0.4em\relax IEEE, 2012, pp. 5026--5033.

\bibitem{brax2021github}
\BIBentryALTinterwordspacing
C.~D. Freeman, E.~Frey, A.~Raichuk, S.~Girgin, I.~Mordatch, and O.~Bachem, ``Brax - a differentiable physics engine for large scale rigid body simulation,'' 2021. [Online]. Available: \url{http://github.com/google/brax}
\BIBentrySTDinterwordspacing

\bibitem{makoviychuk2021isaac}
V.~Makoviychuk, L.~Wawrzyniak, Y.~Guo, M.~Lu, K.~Storey, M.~Macklin, D.~Hoeller, N.~Rudin, A.~Allshire, A.~Handa, and G.~State, ``Isaac gym: High performance gpu-based physics simulation for robot learning,'' 2021.

\bibitem{fu2021minimizing}
Z.~Fu, A.~Kumar, J.~Malik, and D.~Pathak, ``Minimizing energy consumption leads to the emergence of gaits in legged robots,'' \emph{arXiv preprint arXiv:2111.01674}, 2021.

\bibitem{bellegarda2022robust}
G.~Bellegarda, Y.~Chen, Z.~Liu, and Q.~Nguyen, ``Robust high-speed running for quadruped robots via deep reinforcement learning,'' in \emph{2022 IEEE/RSJ International Conference on Intelligent Robots and Systems (IROS)}.\hskip 1em plus 0.5em minus 0.4em\relax IEEE, 2022, pp. 10\,364--10\,370.

\bibitem{duan2023learning}
H.~Duan, B.~Pandit, M.~S. Gadde, B.~J. van Marum, J.~Dao, C.~Kim, and A.~Fern, ``Learning vision-based bipedal locomotion for challenging terrain,'' \emph{arXiv preprint arXiv:2309.14594}, 2023.

\bibitem{grimminger2020open}
F.~Grimminger, A.~Meduri, M.~Khadiv, J.~Viereck, M.~W{\"u}thrich, M.~Naveau, V.~Berenz, S.~Heim, F.~Widmaier, T.~Flayols \emph{et~al.}, ``An open torque-controlled modular robot architecture for legged locomotion research,'' \emph{IEEE Robotics and Automation Letters}, vol.~5, no.~2, pp. 3650--3657, 2020.

\bibitem{jallet2023proxddp}
W.~Jallet, A.~Bambade, E.~Arlaud, S.~El-Kazdadi, N.~Mansard, and J.~Carpentier, ``Proxddp: Proximal constrained trajectory optimization,'' 2023.

\bibitem{osqp}
\BIBentryALTinterwordspacing
B.~Stellato, G.~Banjac, P.~Goulart, A.~Bemporad, and S.~Boyd, ``{OSQP}: an operator splitting solver for quadratic programs,'' \emph{Mathematical Programming Computation}, vol.~12, no.~4, pp. 637--672, 2020. [Online]. Available: \url{https://doi.org/10.1007/s12532-020-00179-2}
\BIBentrySTDinterwordspacing

\bibitem{tonneau2020sl1m}
S.~Tonneau, D.~Song, P.~Fernbach, N.~Mansard, M.~Ta{\"\i}x, and A.~Del~Prete, ``Sl1m: Sparse l1-norm minimization for contact planning on uneven terrain,'' in \emph{2020 IEEE International Conference on Robotics and Automation (ICRA)}.\hskip 1em plus 0.5em minus 0.4em\relax IEEE, 2020, pp. 6604--6610.

\bibitem{haarnoja2018soft}
T.~Haarnoja, A.~Zhou, P.~Abbeel, and S.~Levine, ``Soft actor-critic: Off-policy maximum entropy deep reinforcement learning with a stochastic actor,'' in \emph{International conference on machine learning}.\hskip 1em plus 0.5em minus 0.4em\relax PMLR, 2018, pp. 1861--1870.

\bibitem{yang2022safe}
T.-Y. Yang, T.~Zhang, L.~Luu, S.~Ha, J.~Tan, and W.~Yu, ``Safe reinforcement learning for legged locomotion,'' in \emph{2022 IEEE/RSJ International Conference on Intelligent Robots and Systems (IROS)}.\hskip 1em plus 0.5em minus 0.4em\relax IEEE, 2022, pp. 2454--2461.

\bibitem{he2024agile}
T.~He, C.~Zhang, W.~Xiao, G.~He, C.~Liu, and G.~Shi, ``Agile but safe: Learning collision-free high-speed legged locomotion,'' in \emph{arXiv}, 2024.

\bibitem{alshiekh2018safe}
M.~Alshiekh, R.~Bloem, R.~Ehlers, B.~K{\"o}nighofer, S.~Niekum, and U.~Topcu, ``Safe reinforcement learning via shielding,'' in \emph{Proceedings of the AAAI conference on artificial intelligence}, vol.~32, no.~1, 2018.

\bibitem{fan2024learn}
K.~Fan, Z.~Chen, G.~Ferrigno, and E.~De~Momi, ``Learn from safe experience: Safe reinforcement learning for task automation of surgical robot,'' \emph{IEEE Transactions on Artificial Intelligence}, 2024.

\bibitem{chow2018risk}
Y.~Chow, M.~Ghavamzadeh, L.~Janson, and M.~Pavone, ``Risk-constrained reinforcement learning with percentile risk criteria,'' \emph{Journal of Machine Learning Research}, 2018.

\bibitem{tessler2018reward}
C.~Tessler, D.~J. Mankowitz, and S.~Mannor, ``Reward constrained policy optimization,'' \emph{arXiv preprint arXiv:1805.11074}, 2018.

\bibitem{achiam2017constrained}
J.~Achiam, D.~Held, A.~Tamar, and P.~Abbeel, ``Constrained policy optimization,'' in \emph{International conference on machine learning}.\hskip 1em plus 0.5em minus 0.4em\relax PMLR, 2017, pp. 22--31.

\bibitem{liu2020ipo}
Y.~Liu, J.~Ding, and X.~Liu, ``Ipo: Interior-point policy optimization under constraints,'' in \emph{Proceedings of the AAAI conference on artificial intelligence}, vol.~34, no.~04, 2020, pp. 4940--4947.

\bibitem{zhang2022penalized}
L.~Zhang, L.~Shen, L.~Yang, S.~Chen, B.~Yuan, X.~Wang, and D.~Tao, ``Penalized proximal policy optimization for safe reinforcement learning,'' \emph{arXiv preprint arXiv:2205.11814}, 2022.

\bibitem{sun2022constrained}
H.~Sun, Z.~Xu, Z.~Peng, M.~Fang, T.~Wang, B.~Dai, and B.~Zhou, ``Constrained mdps can be solved by eearly-termination with recurrent models,'' in \emph{NeurIPS 2022 Foundation Models for Decision Making Workshop}, 2022.

\bibitem{charnes1959chance}
A.~Charnes and W.~W. Cooper, ``Chance-constrained programming,'' \emph{Management science}, vol.~6, no.~1, pp. 73--79, 1959.

\bibitem{nemirovski2007convex}
A.~Nemirovski and A.~Shapiro, ``Convex approximations of chance constrained programs,'' \emph{SIAM Journal on Optimization}, vol.~17, no.~4, pp. 969--996, 2007.

\bibitem{rl-games2021}
D.~Makoviichuk and V.~Makoviychuk, ``rl-games: A high-performance framework for reinforcement learning,'' \url{https://github.com/Denys88/rl\_games}, May 2021.

\end{thebibliography}

\section*{Appendix}

\subsection{Hyperparameters}\label{appendix:hyperparameters}

\cite{rl-games2021} details the meaning of some hyperparameters.

\begin{table}[ht]
\centering
\caption{Environment hyperparameters}\label{table:hyp_env}
\begin{tabular}{c | c }
\toprule
Number of envs. & 4096 \\
Random $v_x$ range & $[-0.3, 1.0]$ m/s \\ 
Random $v_y$ range & $[-0.7, 0.7]$ m/s \\ 
Random $\omega_z$ range & $[-0.78, 0.78]$ rad/s \\ 
Proportional gain & 4.0 Nm/rad \\ 
Derivative gain & 0.2 Nm/(rad/s) \\ 
Action scaling & 0.5 \\ 
Default leg angles & [0.05, 0.4, -0.8] rad \\ 
Simulation time step & 5 ms \\ 
Episode length & 10 s \\ 
Height scan grid & 13 x 11 points \\ 
Height scan step & 8 cm \\ 
\bottomrule
\end{tabular}
\end{table}

\begin{table}[ht]
\centering
\vspace{-3mm}
\caption{Learning hyperparameters}\label{table:hyp_ppo}
\begin{tabular}{c | c }
\toprule
Actor network & [512, 256, 128] \\
Critic network & [512, 256, 128] \\
Activation & Elu \\
Discount factor & 0.99 \\
GAE coefficient & 0.95 \\
PPO clipping & 0.2 \\
Entropy coefficient & 1e-3 \\
Learning rate & 3e-4 \\
Learning rate schedule & Adaptive \\
KL threshold for adaptive schedule & 8e-3 \\
Maximum gradient norm & 1.0 \\
Horizon length & 24 \\
Minibatch size & 16384 \\
Mini epochs & 5 \\
Critic coefficient & 2 \\
\bottomrule
\end{tabular}
\end{table}

\begin{table}[ht]
\centering
\vspace{-3mm}
\caption{Constraints hyperparameters}
\begin{tabular}{c | c }
\toprule
Torque $\tau^{\text{lim}}$  & 3 Nm \\
Joint velocity $\dot q^{\text{lim}}$& 16 rad/s \\
Joint acceleration $\ddot q^{\text{lim}}$ & 800 rad/s${}^2$ \\
Action rate $\dot q^{\text{des, lim}}$ & 80 rad/s \\
Base orientation $base^{\text{lim}}$ & 0.1 rad \\
Contact force $f^{\text{lim}}$ & 50 N \\
Hip angle $hip^{\text{lim}}$ &  0.2 rad \\
Air time $t^{\text{target}}_{\text{air time}}$& 0.25s  \\
Number of foot contacts $n_\text{foot contact}^\text{target}$ & 2 \\
Velocity tracking $\epsilon_\text{track}$ & 0.2 m/s or rad/s \\
\bottomrule
\end{tabular}
\end{table} 

\begin{table}[ht]
\centering
\vspace{-3mm}
\caption{Ranges and dimensions of uniform noise for randomizing the dynamics and observations.}\label{table:noise_cstr}
\begin{tabular}{r | l } 
 \toprule
 \multicolumn{2}{ c }{Random Observations:} \\
 \midrule
 $\theta_{\text{body}}$ & $U^{3}(-0.05, 0.05)$ \\
 $\omega_{\text{body}}$ & $U^{3}(-0.001, 0.001)$  \\ 
 $q$ & $U^{12}(-0.01, 0.01)$ \\
 $\dot{q}$ & $U^{12}(-0.2, 0.2)$ \\
 $h_{\text{scan}}$ & $U^{143}(-0.01, 0.01)$  \\
\end{tabular}
\begin{tabular}{r | l } 
 \toprule
 \multicolumn{2}{ c }{Random Dynamics:} \\
 \midrule
Ground Friction & $ U(0.5, 1.25)$ \\
 \bottomrule
\end{tabular}
\end{table}




\end{document}